\documentclass[letterpaper, 10 pt, conference]{ieeeconf}  

\IEEEoverridecommandlockouts                              

\title{\LARGE \bf A GP-based Robust Motion Planning Framework for Agile Autonomous Robot Navigation and Recovery in Unknown Environments}
\author{Nicholas Mohammad, Jacob Higgins, Nicola Bezzo
\thanks{Nicholas Mohammad, Jacob Higgins, and Nicola Bezzo are with the Department of Electrical and Computer Engineering, University of Virginia, Charlottesville, VA 22903, USA 
        {\tt\small \{nm9ur, jdh4je, nbezzo\}@virginia.edu}}%
}%

\newcommand{\subparagraph}{}

\usepackage{graphicx}
\usepackage{epstopdf}
\usepackage{amsmath}
\usepackage{amssymb}
\usepackage{subfigure}
\usepackage{multirow}
\usepackage{pbox}
\usepackage{algorithm}
\usepackage{algpseudocode}
\usepackage{titlesec}
\usepackage{bm}
\usepackage{url}
\usepackage{dblfloatfix} 
\usepackage{booktabs} 
\usepackage{siunitx}  
\usepackage{float}

\usepackage{algpseudocode,algorithm,algorithmicx}  
\algrenewcommand\algorithmicrequire{\textbf{Precondition:}}  
\algrenewcommand\algorithmicensure{\textbf{Postcondition:}}

\renewcommand{\baselinestretch}{0.93}

\usepackage{xcolor}

\DeclareMathOperator*{\argmin}{arg\,min}

\begin{document}

\graphicspath{ {./figs2/} }
\maketitle

\begin{abstract}
For autonomous mobile robots, uncertainties in the environment and system model can lead to failure in the motion planning pipeline, resulting in potential collisions. In order to achieve a high level of robust autonomy, these robots should be able to proactively predict and recover from such failures. To this end, we propose a Gaussian Process (GP) based model for proactively detecting the risk of future motion planning failure. When this risk exceeds a certain threshold, a recovery behavior is triggered that leverages the same GP model to find a safe state from which the robot may continue towards the goal. The proposed approach is trained in simulation only and can generalize to real world environments on different robotic platforms. Simulations and physical experiments demonstrate that our framework is capable of both predicting planner failures and recovering the robot to states where planner success is likely, all while producing agile motion. 

\vspace{5pt}
\emph{Note: }Videos of the simulations and experiments are provided in the supplementary material and at {\url{https://www.bezzorobotics.com/nm-icra24}}.
\end{abstract}

\section{Introduction} \label{sec:intro}
Robust motion planning for autonomous mobile robots (AMR) remains an open problem for the robotics community. One of the main challenges is to navigate through environments in the presence of uncertainty, like an unknown map a priori or inaccurate system models. For example, this lack of robustness was clearly evidenced at the ICRA BARN challenge \cite{xiao2022barn,mohammad2022occlusion}, in which no team was able to navigate a robot through an unknown, cluttered environment without any collisions\footnote{our team placed second in this competition}. Within the navigation stack, the cause of such runtime failures and possible collisions is typically attributed to the motion planning pipeline.

To prevent such situations, reactive approaches have been developed that detect potentially risky states as they occur \cite{zhang2023adarm}. These reactive approaches, however, suffer from poor performance because they are often tuned to be conservative and overly cautious, since it is better to actively avoid unsafe states before they occur. They also do not perform well for high-inertial systems which need an appropriately large reaction time in order to avoid collision. Alternatively, proactive approaches classify future robot states as safe or unsafe based on the current sensor readings and motion plan \cite{ji2022paad,kahn2021badgr}.
These approaches often rely on complex deep learning models which require exhaustive real world training data to detect the safety of future states. Furthermore, these proactive approaches do not solve the problem of recovery after detecting such risky states in the planning horizon.

\begin{figure}[t!]
\centering
\subfigure[]{\includegraphics[width=0.298\textwidth]{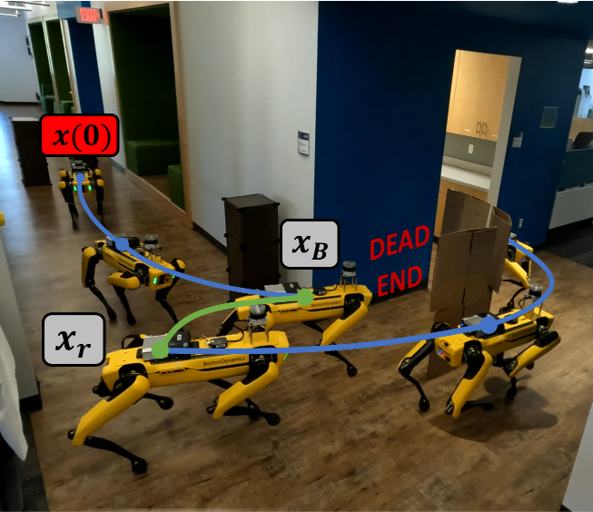}}\label{fig:exp_spot_snapshots}
\subfigure[]{\includegraphics[width=.17\textwidth]{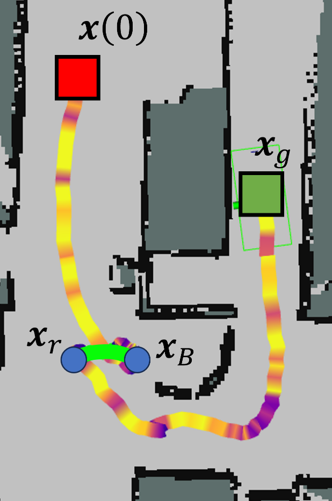}}\label{fig:exp_spot_rviz}
\vspace{-10pt}
\caption{Example in which an unexpected dead end could cause a motion planning failure. The proposed approach predicts the risk of such a failure and recovers before it can occur.}
\label{fig:experiment_spot}
\vspace{-15pt}
\end{figure}

Consider the case in Fig.~\ref{fig:experiment_spot}, where the robot is tasked with navigating quickly through an unknown, occluded environment. Without any proactive scheme, the robot suddenly recognizes the dead end, and does not have enough time to stop before collision.
If these motion planning failures are proactively predicted for future states, then the robot can stop before crashing ($\bm{x}_B$  in Fig.~\ref{fig:experiment_spot}), maneuver to a safe recovery point $\bm{x}_r$, then return to nominal planning towards the final goal. This exact test case will be discussed in Sec.~\ref{sec:experiment}, along with the experimental results.

To achieve this behavior, in this work we propose a proactive- and recovery-focused approach that seeks to predict the risk of failure for a receding horizon, safe corridor motion planner, as well as recover from these potential failures. Such a planner is chosen due to its effectiveness at navigating unknown environments as well as its ubiquity within the robotics community. Additionally, we have found that such a planner requires a relatively small number of features to correctly classify potential failures. A Gaussian Process (GP) is trained on simulated data to predict failures along the planned receding horizon trajectory. When the predicted risk meets a certain threshold criterion, the robot is stopped and a recovery behavior is engaged. This process leverages the same GP to find a nearby safe state from which the robot can safely negotiate its immediate environment and continue motion towards its ultimate goal.

The contribution of this work is a complete and robust motion planning pipeline for robot navigation in unknown environments with two main innovations: 1) \textbf{a proactive planner failure detection scheme} in which a model agnostic, proactive GP-based approach detects and predicts future planning failures and their risk within a horizon, \textit{without the need to retrain between simulation and the real world} and 2) \textbf{a robust recovery scheme} in which a GP-based, sampling-based recovery method  drives the robot to a safe recovery point in order to continue with nominal planning.

\section{Related Work} \label{sec:rel_lit}
While motion planning is an active field of research within the robotics community, the problem of robust, agile navigation through cluttered, unknown environments remains unsolved \cite{xiao2022barn}. 
Many state-of-the-art motion planners impose hard constraints within a nonlinear optimization problem and use numerical solvers to generate the final trajectories within safe corridors \cite{liu2017sfc,gao2018marching,wang2023bspline}. However, random disturbances and occluded obstacles may cause constraint violations at runtime, leading to an inability to generate updated trajectories. \cite{tordesillas2021faster} considers the potential for planner failure by generating an additional safe trajectory which stops within known free space at each planning iteration. However, they do not provide any recovery behaviors in case the vehicle is unable to find feasible trajectories at the stopping point. A popular alternative to the hard-constrained methods are soft-constrained planners, where the hard constraints are converted into differentiable terms and put into the cost function of an unconstrained nonlinear optimization problem \cite{wang2022minco, ren2022bubble}. While the soft-constrained methods generate trajectories even when constraints aren't satisfied, conflicting terms within the cost function can lead to low quality solutions, i.e, unsafe or untrackable trajectories \cite{ning2022optim}. In this paper, we work with the hard-constrained motion planner paradigm and develop an algorithm to monitor for and recover from possible failures proactively at runtime.

Safety monitoring during runtime motion planning is a problem with a catalogue of potential solutions. One well studied technique is Hamilton-Jacobi-Isaacs (HJI) reachability analysis, where safe control is transformed into a formal verification method with
theoretical safety guarantees.
However, HJI reachability requires an accurate model of the system and suffers from the curse of dimensionality \cite{bansal2017hji}. In order to overcome
this problem, recent works have used machine learning techniques to approximate and learn from the generated reachable sets. \cite{devonport2020kriging} leverages Adaptive Kriging using a surrogate GP model and Monte Carlo sampling to approximate the sets at runtime. \cite{yel2020verification} uses a neural network trained on ground truth reachable sets to output binary safe/unsafe classifications for planned trajectories. While these works get around the intractability of runtime reachability analysis, they still rely on specific, accurate system models, limiting their generalizability.

Machine learning methods are also used to monitor vehicle safety outside of the reachability context, stopping the robot when anomalous states are detected. \cite{ji2022paad} and \cite{kahn2021badgr} proactively predict anomalous states which lead to collisions and stop the vehicle before reaching them. However, these works either implement trivial backup and rotate recovery behaviors with no consideration for planning success, or rely on humans to perform the recovery for them.  

Our approach leverages machine learning techniques
to monitor vehicle safety through planner failure detection. Specifically, we train on the distribution of failures over hand-selected input features which enable our approach to be
model agnostic and require only training data from simulation. To the best of our knowledge, our work is the first to utilize a learning component to both proactively predict future planning failures and recover after prediction.

    
\section{Problem Formulation} \label{sec:problem}  

Given a mobile robot system tasked to navigate an unknown environment, let $\dot{\bm{x}}=g(\bm{x},\bm{u})$ define the equations of motion for the system with state $\bm{x}\in\mathbb{R}^{n_x}$ and control inputs $\bm{u}\in\mathbb{R}^{n_u}$. These controls are produced by a low-level controller that is tracking a time-based trajectory $\bm{\tau}(t;t_0)\in\mathbb{R}^{n_x}$ generated at time $t_0$. The purpose of this trajectory is to provide a high-level path plan over a future horizon $t\in[t_0,t_0+T_H]$ from the current state of the robot $\bm{x}(t_0)$ towards a goal $\bm{x}_g$ while avoiding the state subset $\mathcal{X}_O(t_0)$ occupied by obstacles currently known to the robot.
While tracking this trajectory, information about obstacles in the environment are updated at runtime so that, in general, $\mathcal{X}_O(t)\neq \mathcal{X}_O(t_0)$.
This means the trajectory $\bm{\tau}(t;t_0)$ has the potential to collide with these newly discovered obstacles; if this is the case, then a new trajectory must be re-planned. Practical path planners, however, suffer from planning failures within certain situations due to infeasible constraints for the current planning iteration. 
While a single path-planning failure may not be fatal, several failures within the planning horizon could lead to unsafe situations for the robot. 

\textbf{Problem 1: \textit{{Proactive Planner Failure Detection}:}}
Let $\left\{\hat{\bm{x}}_i\right\}$ be a set of predicted future states for the robot while tracking $\bm{\tau}$ over some horizon $T_F \leq T_H$. For a given motion planning policy $\Pi$, define the random variable $Z_{\bm{\tau}}\in\mathbb{W}$ as the number of motion planning failures that occurs from $t_0$ to $t_0+T_F$ while the robot tracks $\bm{\tau}$, with $P(Z_{\bm{\tau}})$ denoting their probabilities.
We seek the creation of a risk metric $\rho_{\bm{\tau}}\in\mathbb{R}$ that maps from $Z_{\bm{\tau}}$ to a single real number that characterizes the risk of path planner failure over $T_F$. 


\textbf{Problem 2: \textit{Recovery After Failure Detection: }} We seek a recovery strategy $\Pi_r(\bm{x}_0)$ that, when the risk $\rho_{\bm{\tau}}$ exceeds a threshold $\psi_\rho$, stops the robot and performs a recovery behavior to reduce the risk of planner failure back down to an acceptable level. Specifically, define $Z_{\bm{x}}\in\{0,1\}$ to represent the success ($0$) or failure ($1$) of $\Pi$ from state $\bm{x}$. The objective of the recovery policy $\Pi_r$ is to locate and control the vehicle to a nearby state $\bm{x}_r$ which maximizes the expected success of $\Pi$:
\begin{equation}
    \bm{x}_r = \argmin_{\bm{x} \not\in \mathcal{X}_O} \mathbb{E}\left[Z_{\bm{x}}\right].
\end{equation}

In the following section, we discuss in detail the design of $\Pi$ and $\Pi_r$, and demonstrate that proactively detecting planner failure and recovering after detection can be achieved by leveraging the same data-informed model. 





\section{Approach}\label{sec:approach}
\begin{figure*}[ht!]
    \centering
    \includegraphics[width=.98\textwidth]{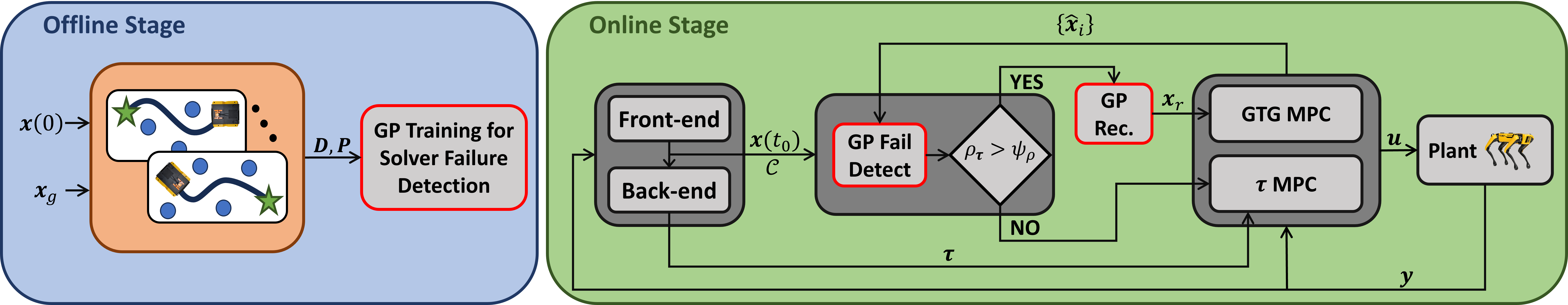}
    \caption{Block diagram for the proposed approach. }
    \label{fig:approach_diagram}
    \vspace{-5pt}
\end{figure*}
\begin{figure*}[ht!]
    \centering
    \subfigure[front-end]{\includegraphics[width=0.232\textwidth]{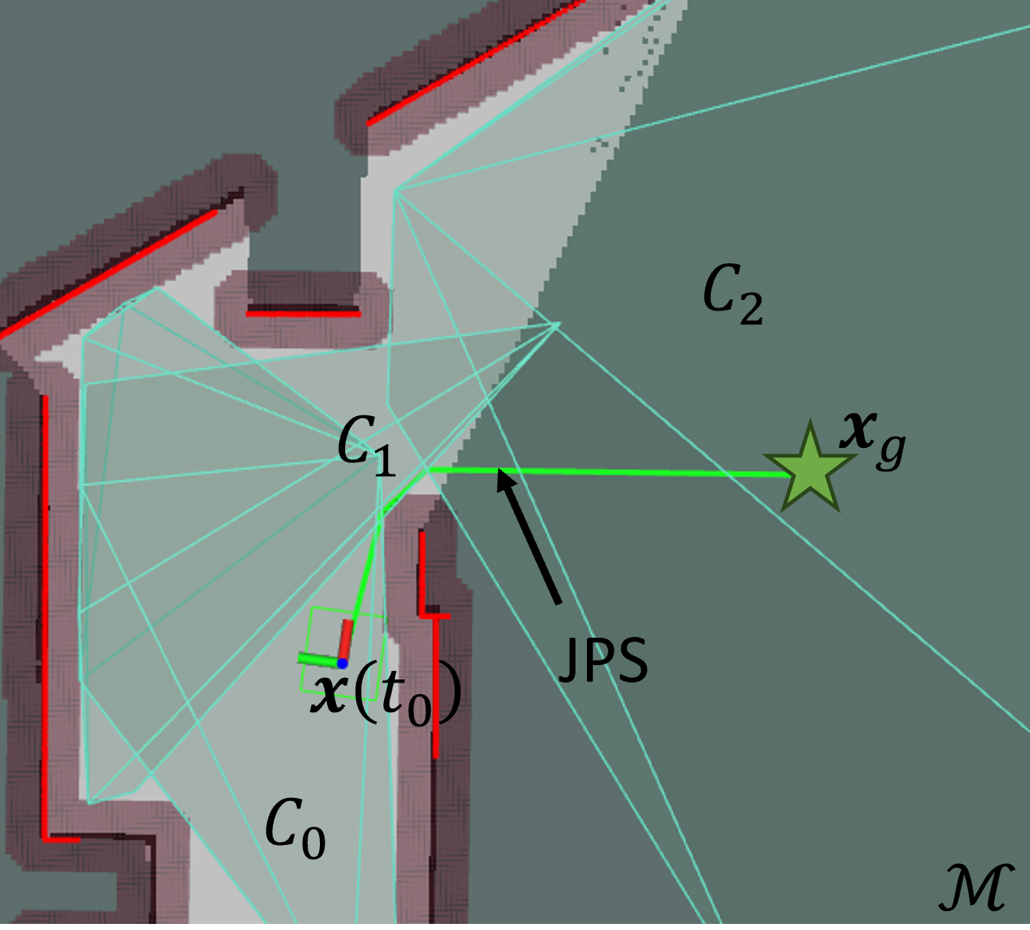}\label{fig:front_end}} 
	\subfigure[back-end]{\includegraphics[width=0.232\textwidth]{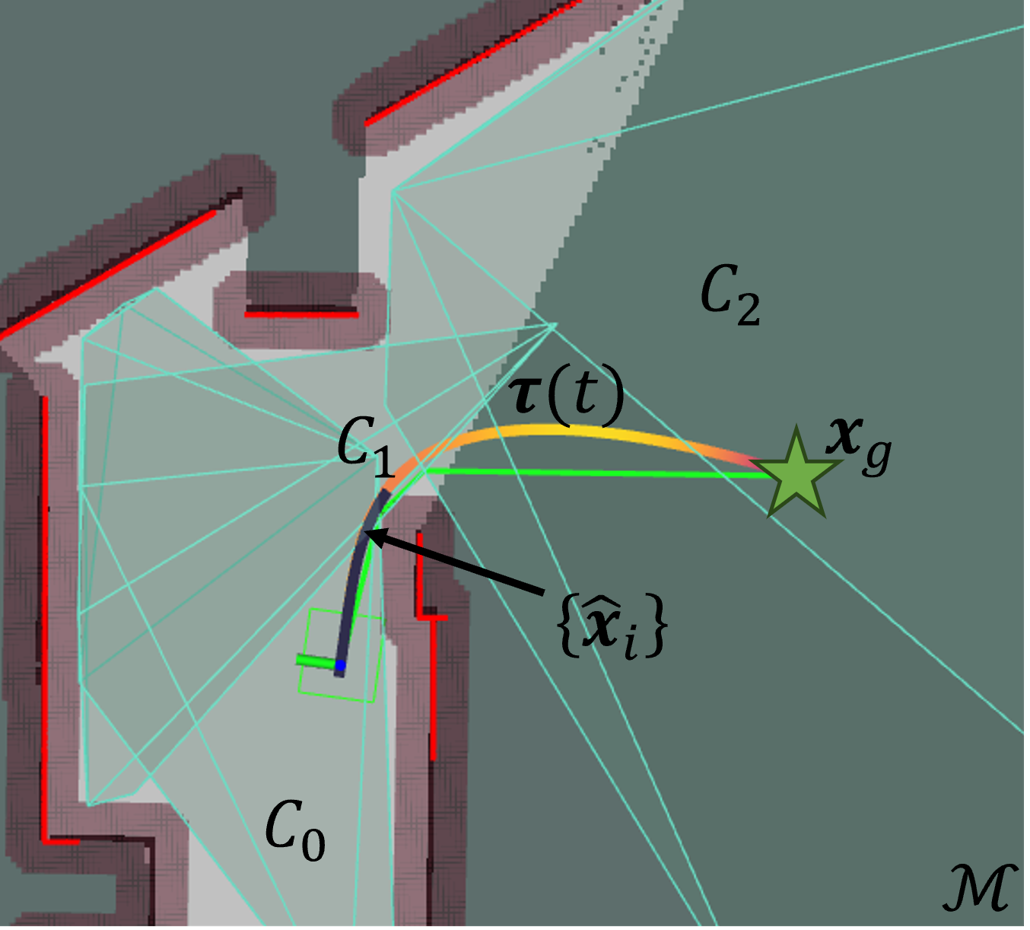}\label{fig:back_end}}
	\subfigure[failure detection]{\includegraphics[width=0.237\textwidth]{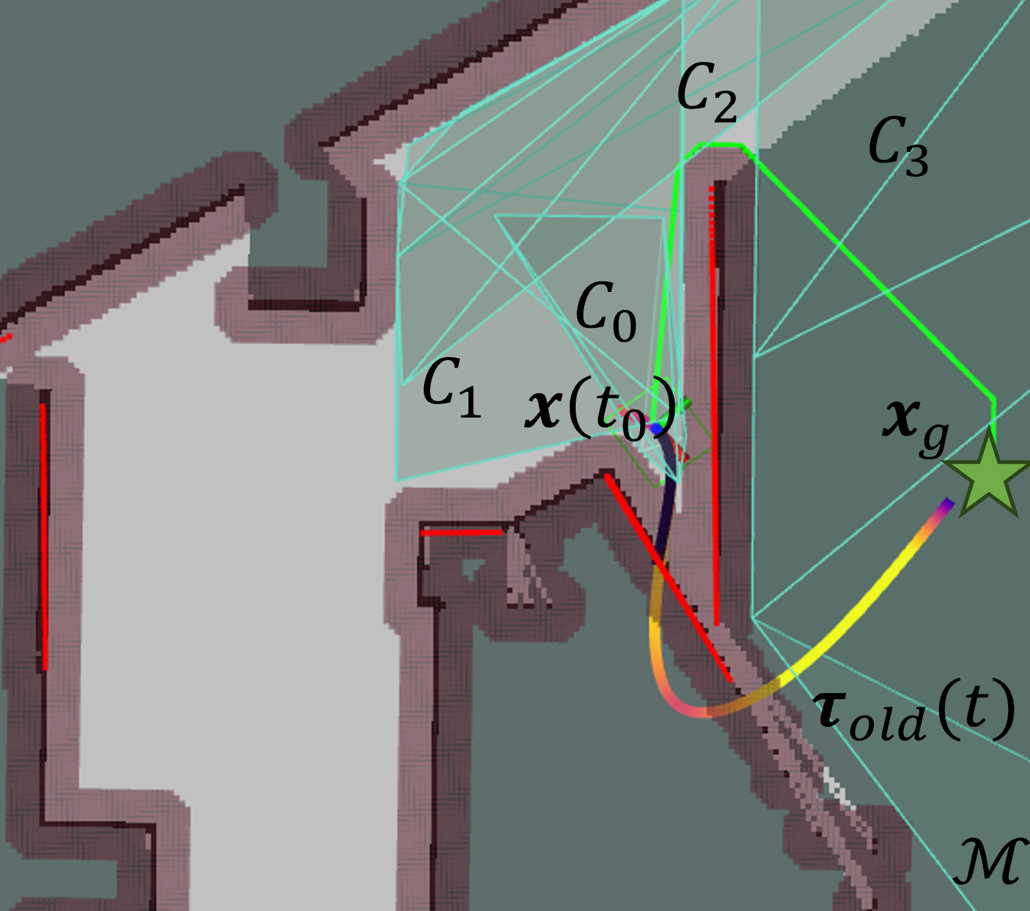}\label{fig:fail_detect}}
	\subfigure[recovery]{\includegraphics[width=0.234\textwidth]{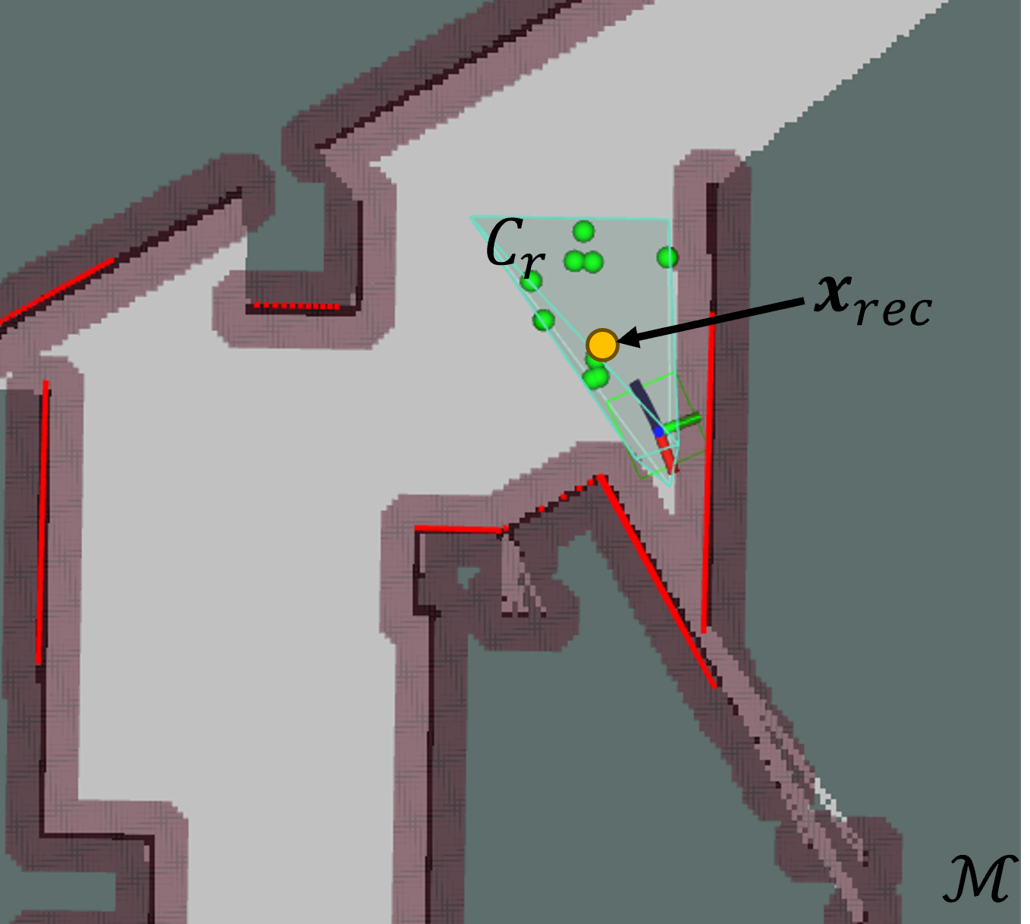}\label{fig:recovery_execute}}
    \caption{An example case study investigated in this work visualizing the complete navigation pipeline (a,b), in which the receding horizon, safe corridor motion planner fails (c), prompting the recovery pipeline to take over and recover the system (d). For $\bm{\tau}(t)$, brighter colors denote higher speeds.}
	\label{fig:planner_demonstration}
    \vspace{-15pt}
\end{figure*}
We propose a GP regression-based scheme to assess the risk of future motion planning failure while tracking a trajectory $\bm{\tau}(t)$. Data were collected from simulations that record motion planning successes and failures in various states that the robot may encounter during typical operation. This data were used to train a GP regression model to predict the probability of motion planning failure for individual states over a future horizon. Fig.~\ref{fig:approach_diagram} shows the outline of our approach. The front-end of the motion planner policy $\Pi$ generates a corridor $\mathcal{C}$ of convex polytopes, illustrated in Fig.~\ref{fig:front_end}. The corridor is then sent to the back-end for final trajectory generation $\bm{\tau}(t)$ (see Fig.~\ref{fig:back_end}). A Model Predictive Controller (MPC) is then used to generate the control signal to track $\bm{\tau}(t)$, generating a sequence of future robot states $\{\hat{\bm{x}}_i\}$ over horizon $T_F$. These states, along with the corridor $\mathcal{C}$, are used to predict the risk of motion planning failure $\rho_{\bm{\tau}}$. Consider the case shown in Fig.~\ref{fig:fail_detect}, where the vehicle is driving towards a previously occluded dead-end. If the predicted risk from our GP-based failure detection model exceeds a user-defined threshold $\psi_\rho$, then the recovery behavior is triggered, and a recovery goal $\bm{x}_r$ is sent to our go-to-goal (GTG) MPC to bring the vehicle to a state where solver success is likely (Fig.~\ref{fig:recovery_execute}).
In the following sections, we describe in detail our motion planner failure prediction and recovery framework, starting with a brief background of the base motion planner.

\subsection{Motion Planner Preliminaries} \label{subsec:preliminaries}

1) \textit{Planner Front-End.} The front-end starts with the global occupancy map $\mathcal{M}$, which is generated by fusing data from an onboard depth sensor, along with the current state of the vehicle, $\bm{x}(t_0)$, and the goal state $\bm{x}_g$. As shown in Fig.~\ref{fig:front_end}, an initial 0-order path within the free and unknown space of $\mathcal{M}$ is generated by using a graph-based, global planner. In this work, we use the Jump Point Search (JPS) algorithm \cite{harabor2011jps}, due to the reduced computational complexity when compared to other common algorithms like A$^*$ . 

A corridor $\mathcal{C}$ of intersecting convex polytopes is then established along this generated initial path, in order to connect $\bm{x}(t_0)$ to $\bm{x}_g$. Each $C_i \in \mathcal{C}$ is represented as an H-polytope defined by a matrix $\bm{A}_i$ and vector $\bm{b}_i$ that define a convex set of points $\bm{p}\in\mathbb{R}^2$ in the $xy$ plane
\begin{equation}
    C_i = \{\bm{p} \in \mathbb{R}^2 \, | \, \bm{A_i}\bm{p} \leq \bm{b_i} \}. \label{eq:h_poly}
\end{equation} 

In order to generate each $C_i$ of the corridor $\mathcal{C}$, we rely on the gradient-based optimization approach in \cite{wang2022minco}. 
With $\mathcal{C}$ constructed, the corridor is sent along with $\bm{x}(t_0)$ and $\bm{x}_g$ to the back-end optimization to find the final trajectory $\bm{\tau}(t)$.

2) \textit{Planner Back-End.} We represent the trajectory $\bm{\tau}(t)$ (shown in Fig.~\ref{fig:back_end}) as a collection of $N$ cubic ($n=3$) B\'ezier curves. We use these curves for the trajectory formulation as they are a commonly utilized basis with several salient properties for corridor-based motion planners \cite{gao2018marching}. One useful property of the B\'ezier curve $\bm{\tau}_j(t)$ is that it is fully contained within the simplex formed by the control points $\bm{q}^i_j, \, i \in [0,n]$. Thus, for $\bm{\tau}_j(t)$ to be contained within a convex polytope $C$, it is sufficient to ensure that $\bm{q}_j^i \in C, \, \forall i \in [0,n]$. 
To generate the final trajectory, we leverage the FASTER solver \cite{tordesillas2021faster}, altered to convert the B\'ezier control points of each trajectory segment $\bm{\tau}_j(t)$ into the MINVO basis \cite{tordesillas2022minvo} during optimization to improve solver success rate. Once $\bm{\tau}(t)$ has been found, it is sent to the tracking MPC to be executed on the robot.


\subsection{Failure Modes: Front-End vs Back-End} \label{subsec:failure_modes}
There are two distinct failure modes of the motion planner described in Sec.~\ref{subsec:preliminaries}, both of which will result in $\Pi=\emptyset$: (i) a front-end failure, in which an intersecting corridor $\mathcal{C}$ between $\bm{x}(t_0)$ and $\bm{x}_g$ cannot be found, or (ii) a back-end failure, in which the numerical solver fails to generate a trajectory along the JPS search path. Front-end failures can occur when a feasible search path doesn't exist (e.g., either $\bm{x}(t_0)$ or $\bm{x}_g$ overlap occupied space within $\mathcal{M}$), or when parameters of the JPS are poorly conditioned for generating a corridor $\mathcal{C}$ (e.g., $|\mathcal{C}|$ is high because the planning horizon distance is too large). The front-end of the motion planner implemented in Sec.~\ref{subsec:preliminaries} typically runs in $< 1$ms, thus for a given state $\bm{x}$ and map $\mathcal{M}$, front-end failures are easily determined by simply running the JPS and corridor generation at that state. 

Much more difficult to predict, however, are failures at the back-end of the motion planner due to the fact that the environment is unknown a priori and the optimization is based only on current observations in $\mathcal{M}$. Since the back-end is based on a nonlinear optimizer, it can be difficult to characterize success or failure prior to actually running the back-end solver. Additionally, the time to run the back-end is typically $>100$ms, which is too large to directly test multiple future points for failure. Fig.~\ref{fig:fail_detect} shows an example back-end failure, in which the discovery of a previously unknown wall (shown as undiscovered space in Fig.~\ref{fig:back_end}) requires a new avoidant trajectory to be generated. While the front-end is able to generate a corridor $\mathcal{C}$, the back-end is unable to find a feasible trajectory. 

To concretely define these ideas, let $Z_{\bm{x},\mathcal{C}}\in\left\{0,1\right\}$ represent a success ($0$) or failure ($1$) of the motion planner pipeline, with $Z^f\in\left\{ 0,1 \right\}$ representing a front-end failure and $Z^b\in\left\{0,1\right\}$ representing a back-end failure. Success of the back-end is dependent on success at the front-end, and failure of the front-end is interpreted as a failure of the back-end as well, so that $P\left(Z^b=1|Z^f=1\right)=1$. The probability of entire pipeline failure can be written as
\begin{equation}\label{eq:prob_total_failure}
    P(Z_{\bm{x},\mathcal{C}}) = P\left(Z^b|Z^f\right)P\left(Z^f\right).
\end{equation}
The probability of front-end failure is easily and rapidly checked by running the JPS for a given $\bm{x}$ and $\mathcal{C}$, so that effectively $P(Z^f)\in\left\{0,1\right\}$. Our contribution is in estimating the probability of back-end failure after a front-end success, $P(Z^b|Z^f=0)$. For simplicity in notation, in the rest of the paper we will write this probability as $P(Z^b)$ and drop the dependence on the front-end outcome.

\subsection{Gaussian Process for Predicting Back-End Failure} \label{subsec:gp_backend}
To accurately predict back-end failures, we propose a GP-based regression model trained on statistics inferred from simulated data. We choose GPs due to their non-parametric form and ability to accurately infer from a small dataset. These data relate the robot and map state to the probability of back-end failure $P(Z^b_{\bm{x},\mathcal{C}})$. A GP model $\hat{P}(Z^b_{\bm{x},\mathcal{C}}|\cdot)$ can be trained to predict back-end failure probability at run time over future states $\{\hat{\bm{x}}_i\}$.
These probabilities can then be used to assess the risk of future motion planning failure $\rho_{\bm{\tau}}$ over the entire prediction horizon.

A navigation stack comprising of both the planning policy $\Pi$ and the MPC can be deployed in simulation to gather training examples for the GP model. To generate the training dataset, $\bm{D}$, we use the Poisson random forest dataset from \cite{oleynikova2016forest}, which contains 10 forest worlds, each with a collection of 90 start and goal positions for navigation. A Clearpath Jackal UGV was then tasked to navigate through the worlds in each of the start and goal configurations, collecting back-end success and failure data points at each planning iteration. With these data collected, features which correlate with back-end failure can be found. To promote generality, the chosen features should only depend on the corridor set $\mathcal{C}$, regardless of the sensing modality used to generate it (LiDAR, RGBd, etc.), along with the robot position and its time derivatives, which are common state features for most AMR.

\indent1) \textit{Feature Selection.} Each training tuple contains three pieces of information: (i) robot state $\bm{x}$, (ii) corridor $\mathcal{C}$, and (iii) binary variable $Z^b_{\bm{x},\mathcal{C}}$ which encodes a success or failure of the back-end. With these data, statistical inferences can be made that relate the robot state and corridor to the probability of back-end failure $P(Z^b_{\bm{x},\mathcal{C}})$. 
Through study of various possible features that could be used, we found two which were particularly well-suited to predicting the probability of back-end failure: the minimum time-to-intersect (TTI), $t_C$, from robot state $\bm{x}$ to corridor $\mathcal{C}$, and the number of polytopes that define the corridor $|\mathcal{C}|$. 

The minimum TTI can be found by using the $xy$ position $\bm{p}\in\mathbb{R}^2$ and velocity $\bm{v}\in\mathbb{R}^2$ of the robot state $\bm{x}$, then using kinematic equations to find the minimum TTI of the hyperplanes that define the polytope~$C$ containing $\bm{p}(t_0)$. Formally, if row $\bm{r}_i \in \bm{A}$ and $b_i \in \bm{b}$ form a hyperplane $\bm{r}_i\cdot \langle x,y \rangle = b_i $ of polytope $C$,
then the time to intersect the hyperplane $t_H$ can be calculated as:
\begin{equation}
t_H\left(\bm{r}_i, b_i, \bm{x}\right) =
    \begin{cases}
        \frac{b_i-\bm{r}_i\cdot\bm{p}}{\bm{r}_i\cdot\bm{v}} & \text{if } \bm{r}_i \cdot \bm{v} > 0\\
        \gamma_t & \text{otherwise}
    \end{cases}
\end{equation}
where $\gamma_t$ is a user-defined maximum value for $t_H$ when the vehicle is stationary or moving away from the hyperplane. 
With $t_H$, $t_C$ is calculated as the minimum TTI to the hyperplanes of $C$:
\begin{equation}
    t_C = \min_i \{t_H\left(\bm{r}_i, b_i, \bm{x}\right)\}.
\end{equation}

One of the biggest factors that affect the ability of the back-end solver to find a feasible solution is how close the current robot position $\bm{p}(t_0)$ is located to the boundary of the feasible set $\mathcal{C}$. Intuitively, TTI is an effective predictor of back-end failure because it captures several factors that determine success: (i) The physical distance between $\bm{p}(t_0)$ and the free space boundary, (ii) the velocity of the robot $\bm{v}(t_0)$, and (iii) the heading of the robot. 

\begin{figure}[ht!]
\centering
\subfigure[]{\includegraphics[width=0.23\textwidth]{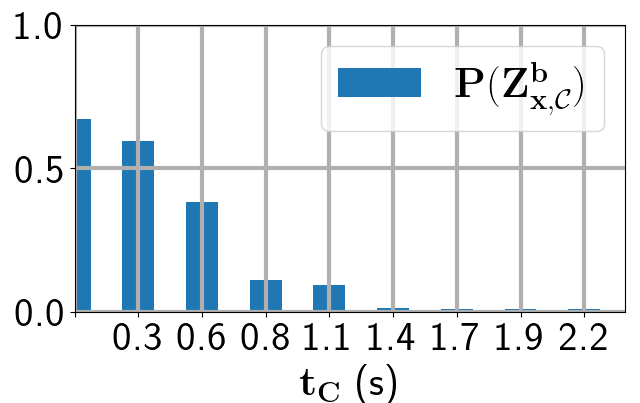}}\label{fig:tti_vs_failure}
\subfigure[]{\includegraphics[width=.23\textwidth]{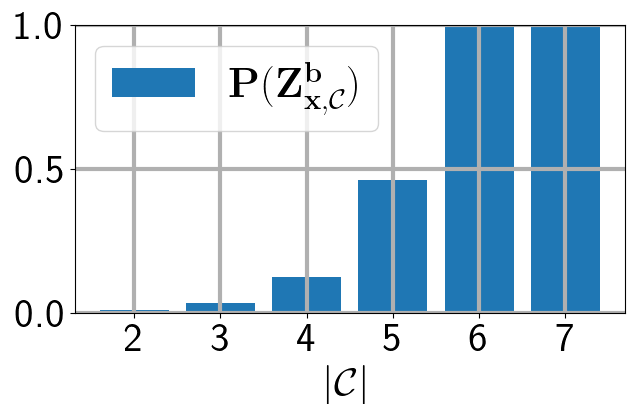}}\label{fig:corridor_len_vs_failure}\\
\vspace{-10pt}
\subfigure[]{\includegraphics[width=.23\textwidth]{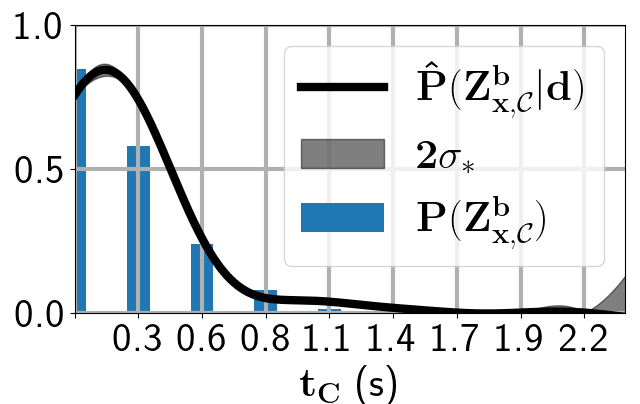}}\label{fig:corridor_2_dist}
\subfigure[]{\includegraphics[width=.23\textwidth]{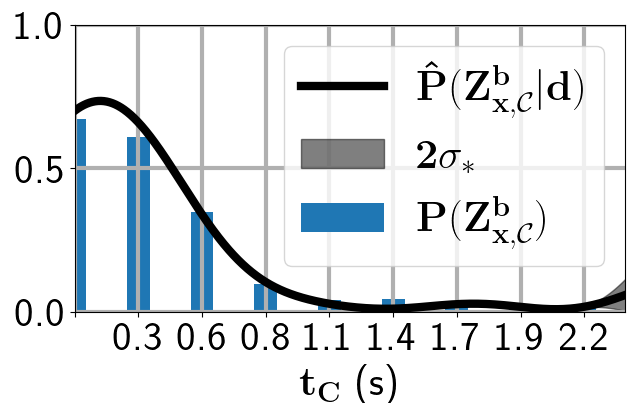}}\label{fig:corridor_3_dist}
\vspace{-8pt}
\caption{Solver failure trends for (a) $t_C$ and $|\mathcal{C}|$ along with learned distribution vs test distribution for (c) $|\mathcal{C}|=2$ and (d) $|\mathcal{C}|=3$.}
\vspace{-5pt}
\label{fig:tti_data}
\end{figure}

In addition to TTI, the cardinality $|\mathcal{C}|$ of the corridor also plays a role in failure of the back-end solver: if $\mathcal{C}$ is defined by many polytopes, then obstacles in the environment necessitate a very non-direct path to be planned for the robot, further complicating the search for a feasible path. 
Together, these two features were used inside a feature vector $\bm{d}(\mathcal{C},\bm{x})=\left[t_C,|\mathcal{C}|\right]$ to infer the probability of back-end failure. To find this probability, the back-end failure training data $\left\{Z^b_{\bm{x},\mathcal{C}}\right\}$ were binned based on feature vector value $\bm{d}$, and ground-truth probability of failure $P(Z^b_{\bm{x},\mathcal{C}})$ was found within each bin.
To validate the choice of input features for training, we plot the probability of back-end failure $P(Z^b_{\bm{x},\mathcal{C}})$ over $t_C$ and $|\mathcal{C}|$, where the correlations are clearly seen in Figs.~\ref{fig:tti_data}(a) and (b). As $t_C$ decreases, the probability of back-end failure increases. Furthermore, as the corridor length $|\mathcal{C}|$ increases, the  probability of failure also increases. 

2) \textit{GP Regression.} The underlying GP model input is defined by a collection of $M$ input training features, $\bm{D}=\left[\bm{d}_0, \dots, \bm{d}_M\right]$, and values $\bm{P}=\left[P_0, \dots, P_M\right]$, with an output defined by a joint Gaussian distribution  \cite{rasmussen2005gaussian}:
\begin{equation}
    \begin{bmatrix} P \\ \hat{P} \end{bmatrix} \sim \mathcal{N}\left( \begin{bmatrix} \mu(\bm{d}) \\ \mu(\bm{d}_*) \end{bmatrix}, \begin{bmatrix} \bm{K} & \bm{K_*} \\ \bm{K}_*^T & \bm{K_{**}} \end{bmatrix} \right), \label{eq:gp}
\end{equation}
where $\bm{K}=\kappa(\bm{D}, \bm{D})$, $\bm{K_*}=\kappa(\bm{D},\bm{D}_*)$ and $\bm{K_{**}}=\kappa(\bm{D}_*, \bm{D}_*)$, $\mu$ is the mean function, $\bm{D}_*$ is the test input, and $\kappa$ is a positive definite kernel function, which is the Radial Basis Function (RBF) in this work. From this, the predictive posterior distribution of $\hat{P}$ given $\bm{D}$ can be expressed as another Gaussian distribution:
\begin{equation}
    \hat{P} \sim \mathcal{N}(\mu_*, \sigma_*^2),
\end{equation}
with $\mu_*$ and $\sigma_*^2$ defined as:
\begin{equation}
    \mu_* = \mu(\bm{D_*}) + \bm{K}_*^T\bm{K}\left(P-\mu(\bm{D})\right)
\end{equation}
\begin{equation}
    \sigma_*^2 = \bm{K}_{**}-\bm{K}_*^T\bm{K}^{-1}\bm{K}_*.
\end{equation}
With this, the estimated probability of back-end failure is taken as the mean values of this posterior:
\begin{equation}
    \hat{P}\left(Z^b_{\bm{x},\mathcal{C}}|\bm{d}\right) = \mu_*.
\end{equation}

To validate the quality of the trained GP models, the distribution of failures over $t_C$ was collected from test worlds outside the forest dataset, and the resulting test set distribution was compared with the learned distribution $\hat{P}(Z^b_{\bm{x},\mathcal{C}}|\bm{d})$ for $|\mathcal{C}|=2$ (Fig.~\ref{fig:tti_data}(c)) and $|\mathcal{C}|=3$ (Fig.~\ref{fig:tti_data}(d)). These plots show the learned distributions closely match the test distribution, demonstrating that the GP models generalize well to new environments.

3) \textit{Defining Planning Risk.} With $\hat{P}(Z^b_{\bm{x},\mathcal{C}}|\bm{d})$ estimating back-end failure, the probability of failure for the entire motion planning pipeline $P(Z_{\bm{x},\mathcal{C}})$ can be calculated using~\eqref{eq:prob_total_failure}. These probabilities can be used to infer the risk of motion planning failure along the future states $\left\{\hat{\bm{x}}_i\right\}$ predicted by the MPC. To formulate this risk, we consider the total number of future motion planning failures $Z_{\bm{\tau}}$ as the salient outcome to track, defined as
\begin{equation}
    Z_{\bm{\tau}} = \sum_{\hat{\bm{x}} \in \{\hat{\bm{x}}_i\}}Z_{\hat{\bm{x}},\mathcal{C}}.
\end{equation}

Because each $Z_{\hat{\bm{x}},\mathcal{C}}$ is a stochastic variable, $Z_{\bm{\tau}}$ is also a stochastic variable. The risk metric chosen in our approach is the expected number of collisions over the future horizon, $\rho_{\bm{\tau}} = \mathbb{E}(Z_{\bm{\tau}})$. The expected value is chosen here for its simplicity and speed to calculate, although other risk metrics may be used as well~\cite{majumdar2020should}. Since each $Z_{\hat{\bm{x}},\mathcal{C}}$ is a Bernoulli random variable with predicted probability $P(Z_{\hat{\bm{x}},\mathcal{C}})$ of failure, the expectation is calculated as:
\begin{equation}
    \rho_{\bm{\tau}} = \sum_{\hat{\bm{x}} \in \{\hat{\bm{x}}_i\}} P(Z_{\hat{\bm{x}},\mathcal{C}}).
\end{equation}

A risk threshold $\psi_\rho$ may be set so that anytime the risk of planner failure over future states $\left\{\hat{\bm{x}}_i\right\}$ exceeds this value, the recovery behavior is triggered.

\subsection{Recovering After Predicted Failures}\label{subsec:recovery}
When $\rho_{\bm{\tau}} > \psi_\rho$ is satisfied, it means that there is a collection of states in the vehicle's future horizon that the planner is likely unable to successfully operate. As such, the vehicle must stop or perform other recovery maneuvers in order to avoid collisions and navigate successfully through said regions. Unlike prior works where human operators intervene to recover the vehicle once failures are detected \cite{ji2022paad}, \cite{kahn2021land}, our framework includes a recovery planner $\Pi_r$ which enables the vehicle to find and execute safe recovery maneuvers autonomously, as illustrated in Fig.~\ref{fig:recovery_execute}.

Once the vehicle has stopped after switching to the recovery mode, the objective is to locate a nearby region where the planner will succeed, i.e., $Z_{\bm{x},\mathcal{C}}=0$. The first step is to sample points uniformly in free space around the current vehicle position $\bm{p}(t_0)$. To do so, an H-polytope $C_{r}$ is generated around $\bm{p}(t_0)$, where hit-and-run Markov-chain Monte Carlo sampling \cite{drake} is used to find $N_p$ candidate positions $\mathcal{P}_{c} = \{\bm{p}_0, \dots, \bm{p}_{N_p}\}$, where $N_p$ is a user-defined parameter. $\mathcal{P}_{c}$ is then converted to states $\mathcal{X}_{c}$ by assuming the vehicle starts from rest. We make this choice because it significantly reduces the sample space and sampling only positions was enough to find recovery states in practice. 
With $\mathcal{X}_{c}$, we find the probability of planner failure, $P(Z_{\bm{x}_i,\mathcal{C}_i})$, at each $\bm{x}_i$, as well as neighboring states in close proximity for consistency. 
If all predictions have failure probability greater than $\eta$, the samples are thrown away and the sampling process is repeated. Here $\eta$ is a user-defined parameter which controls how risk averse the recovery behavior should be. $\bm{x}_r$ is then chosen to be the state with lowest expected failure:
\begin{equation}
    \bm{x}_r = \argmin_{\bm{x}_i \in \mathcal{X}_c} \mathbb{E}\left[Z_{\bm{x_i},\mathcal{C}_i}\right]. \label{eq:rec_state}
\end{equation}

After determining $\bm{x}_r$, the vehicle navigates to the recovery point using the GTG MPC with an added constraint, formulated as in \eqref{eq:h_poly}, where $\bm{p}(t_0)$ must remain in $C_r$ in order to avoid obstacles. Once the vehicle reaches $\bm{x}_r$, the planner switches back to the nominal safe corridor policy $\Pi$ to generate trajectories $\bm{\tau}(t)$ and the entire process repeats.

\section{Simulations}\label{sec:simulation}
Simulations were performed to both train the GP classification model described in \eqref{eq:gp} and validate the proposed approach to detect and recover from motion planning failures. All simulations were performed in Gazebo using Ubuntu 20.04 and ROS Noetic. The robot used in simulation is a Clearpath Robotics Jackal UGV equipped with a $270^\circ$ 2D Lidar depth sensor. Data were collected as described in Sec.~\ref{subsec:gp_backend} and sent to the GP regressions for training.

\begin{figure}[ht!]
\centering
\subfigure[]{\includegraphics[width=.23\textwidth]{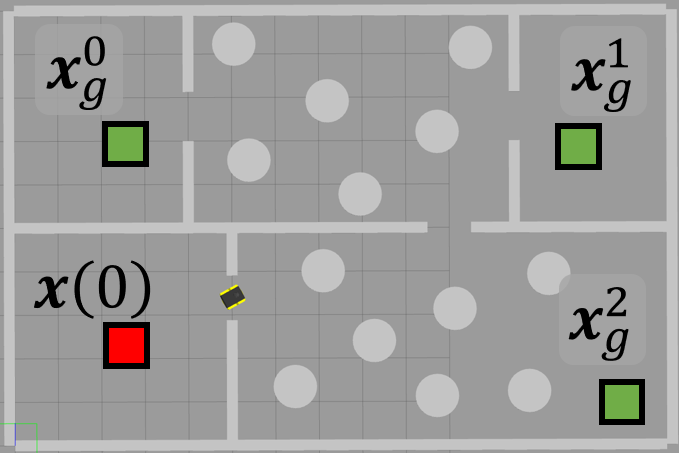}}\label{fig:gazebo_world}
\subfigure[]{\includegraphics[width=.23\textwidth]{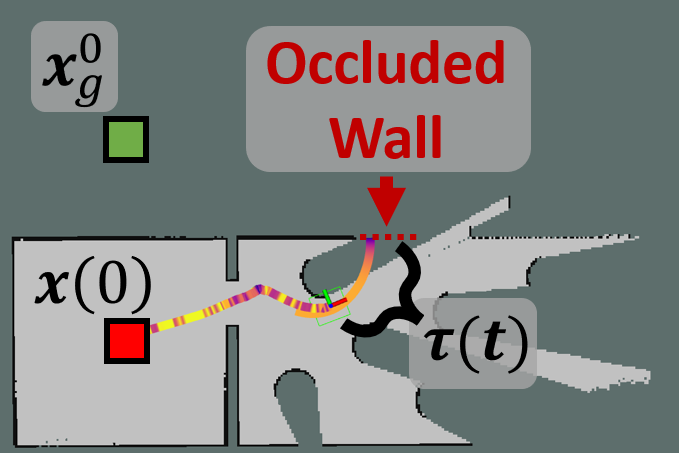}}\label{fig:sim_before_collision}
\quad\quad\quad
\subfigure[]{\includegraphics[width=.235\textwidth]{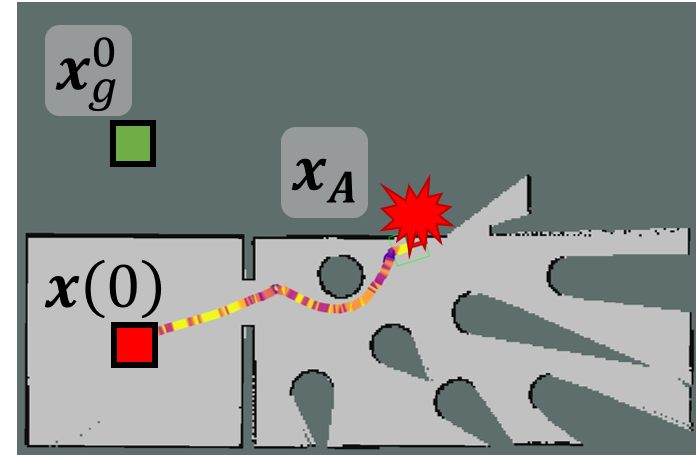}}\label{fig:sim_fail_example}
\subfigure[]{\includegraphics[width=.23\textwidth]{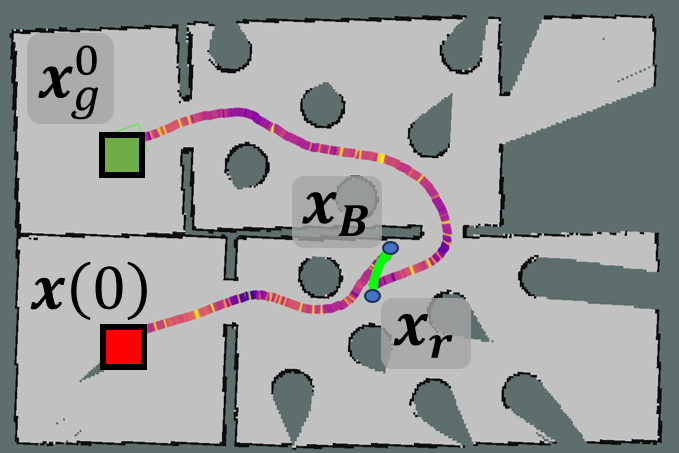}}\label{fig:sim_success_example}
\vspace{-7pt}
\caption{(a) Gazebo world with $1$m doorway and 3 different goals. In (b) an obstacle hides an occluded wall leading to a collision without our framework (c) vs a successful navigation toward $x_g^0$ in (d) with our approach.}
\label{fig:sim_example}
\end{figure}

With the models trained, we then validated our approach in four gazebo worlds of varying difficulty. The base world is a series of connected rooms with either sparse or dense obstacle density and 1m or $2$m wide doorways. In each world we use the same start configuration $\bm{x}(0)$ and three goals $\bm{x}_g^0$, $\bm{x}_g^1$, and $\bm{x}_g^2$. Fig.~\ref{fig:sim_example}(a) shows the world with $1$m doorways and dense obstacle configuration, along with an example navigation failure without our approach (Fig.~\ref{fig:sim_example}(b) and (c)) and success with (Fig.~\ref{fig:sim_example}(d)). In Fig.~\ref{fig:sim_example}(b), the vehicle plans a trajectory $\bm{\tau}(t)$ which intersects a part of the wall occluded by an obstacle. 
Since an avoiding trajectory cannot be computed in time, the vehicle collides with the wall at $\bm{x}_A$ in Fig.~\ref{fig:sim_example}(c). If instead we use our approach, as shown in Fig.~\ref{fig:sim_example}(d),
the robot detects the planner failure proactively and stops at $\bm{x}_B$. A reverse maneuver (green line) is then executed to reach the recovery state $\bm{x}_r$ found using \eqref{eq:rec_state}. The vehicle then switches back to the nominal planner and continues towards $\bm{x}_g^0$.


\begin{figure}[ht!]
    \centering
    \includegraphics[width=.47\textwidth]{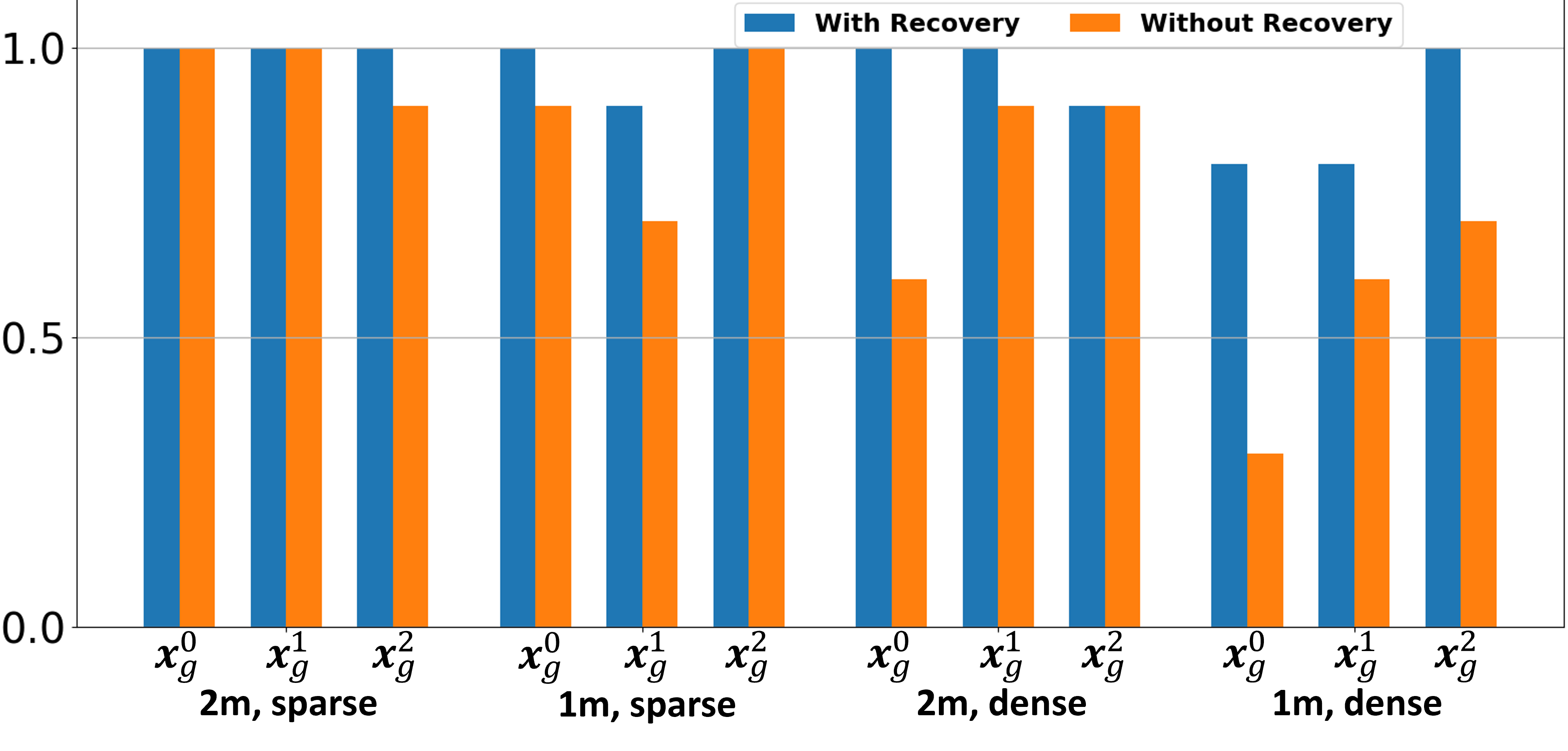}
    \vspace{-8pt}
    \caption{Navigation success rates with (blue) and without (orange) failure detection and recovery for different simulation world and goal combinations.}
    \label{fig:sim_results}
\end{figure}

The remaining $3$ test worlds are generated by varying the doorway width between $1$m and $2$m, as well as varying the obstacle layout between a sparse and dense configuration. For each world tested, the robot is tasked to navigate 10 times to the goals $\bm{x}_g$, creating $30$ test points per world, for $120$ simulations total. The resulting success rates for the motion planner with and without our approach are shown in Fig.~\ref{fig:sim_results} for each goal and world combination, where it can be seen that using our failure detection and recovery framework improves the nominal planner's performance.

  \vspace{-5pt}
\section{Physical Experiments} \label{sec:experiment}

The proposed approach was validated with multiple robots across several experiments, all of which are shown in the supplementary material and website. Presented in this paper are two experiments with two real robotics platforms: a Boston Dynamics Spot quadruped, and the same Jackal differential drive UGV used in simulations. 
For each platform, the same motion planning pipeline was used to generate trajectories $\bm{\tau}$ to follow, using an MPC to generate the control signal $\bm{u}$ to track these trajectories. Lidar sensor readings were provided by Ouster for the Spot, and Velodyne for the Jackal. These were used by the SLAM package Gmapping in order to create a map $\mathcal{M}$ and estimate the state of the robot at run-time as each platform traveled through an environment unknown a priori. To emphasize the generality of the proposed approach, the GP model $\hat{P}(Z_{\bm{x},\mathcal{C}}^b|\bm{d})$ that was used to predict motion planning back-end failures was \textit{trained entirely on data collected in simulation}, demonstrating how the approach is both sensor- and model-agnostic.

\begin{figure}[ht!]
    \centering
    \subfigure[]{\includegraphics[width=0.32\textwidth]{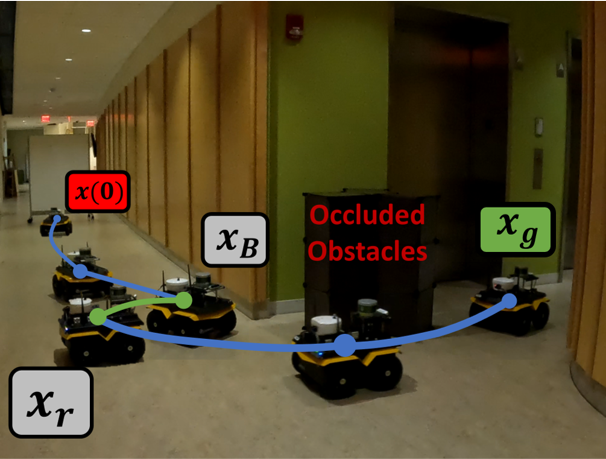}}\label{fig:exp_jackal_snapshots}
    \subfigure[]{\includegraphics[width=.15\textwidth]{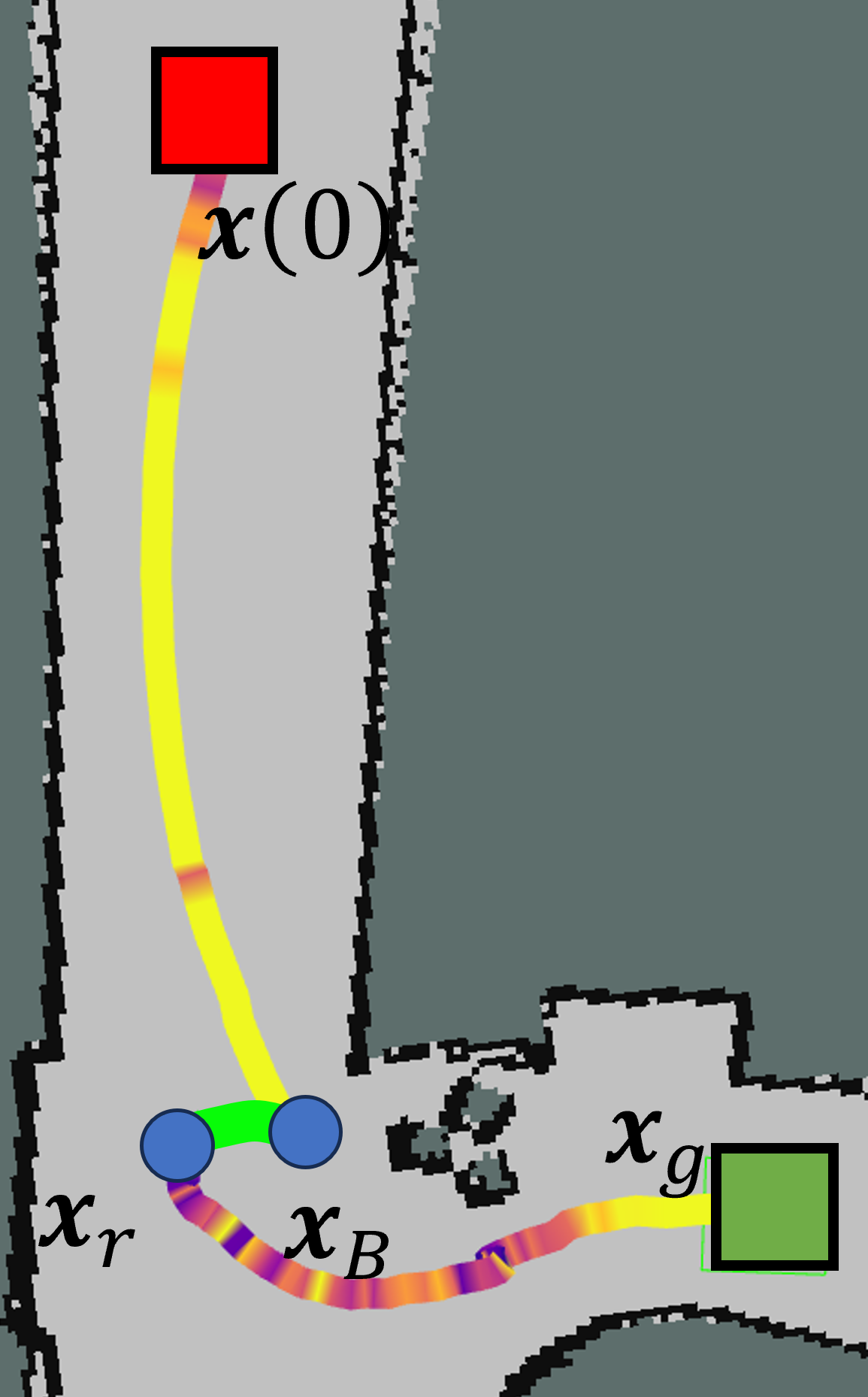}}\label{fig:exp_jackal_rviz}
    \caption{Experiment case in which unexpected occluded obstacles would have caused a motion planner failure without our approach.}
    \label{fig:experiment_jackal}
\end{figure}

Two test cases were setup to test the approach. Fig.~\ref{fig:experiment_jackal} shows the first case in which the Jackal is tasked to move towards a goal around an occluding corner, behind which are occluded obstacles previously unknown to the robot. Fig.~\ref{fig:experiment_spot} shows the second case in which the Spot is tasked with a similar mission, except it must negotiate an unexpected dead-end. 
Without the proposed approach, both cases lead to path-planning failures, which in turn lead to collisions. Both Fig.~\ref{fig:experiment_jackal} and Fig.~\ref{fig:experiment_spot} show snapshots of the proposed approach being used to proactively detect risk of path planning failure $\rho_{\bm{\tau}}$, recovering at $\bm{x}_B$ when $\rho_{\bm{\tau}} > \psi_{\rho}$, moving to a recovery point $\bm{x}_r$, then continuing moving towards $\bm{x}_g$. For these experiments, the risk threshold was $\psi_\rho=3$ {\em expected failures} over the predicted MPC future trajectory.
\vspace{-0pt}


\section{Conclusions and Future Work} \label{sec:conclusion} 
In this work, we have presented a novel GP-based, proactive failure detection and recovery scheme to prevent a mobile robot system from colliding with obstacles. Our approach is shown to improve the performance over a traditional safe corridor motion planner in both simulation and experimental case studies. Furthermore, our approach is model- and sensor-agnostic and can be applied without prior real-world training data due to the careful selection of features.

Future work aims to enhance the system by incorporating distributional learning for failure detection, eliminating the need for multiple GP regressions. Additionally, we would like to utilize this approach for planner switching within a Simplex Architecture and incorporate dynamic obstacles.

\vspace{-0pt}

\section{Acknowledgements}
\vspace{-2pt}
Funding for this research are provided by an Amazon Research Award and by CoStar group.
\vspace{-6pt}
\newpage


\bibliographystyle{IEEEtran}
\bstctlcite{IEEEexample:BSTcontrol}
\bibliography{library}

\end{document}